\documentclass[10pt]{article} 
\usepackage[preprint]{tmlr}

\usepackage{hyperref}
\usepackage{url}

\usepackage{amssymb}
\usepackage{amsmath}

\usepackage{algorithm}
\usepackage{algpseudocode}

\usepackage{booktabs}
\usepackage{multirow}

\usepackage{graphicx}
\graphicspath{ {./images/} }
\usepackage{caption}
\usepackage{subcaption}

\usepackage{color}

\title{Bayesian Quadrature for Neural Ensemble Search}

\author{\name Saad Hamid \email saad@robots.ox.ac.uk \\
\addr University of Oxford
\AND
\name Xingchen Wan \email xwan@robots.ox.ac.uk \\
\addr University of Oxford
\AND
\name Martin J{\o}rgensen \email martinj@robots.ox.ac.uk\\
\addr University of Oxford
\AND
\name Binxin Ru \email robin@robots.ox.ac.uk \\
\addr University of Oxford
\AND
\name Michael Osborne \email mosb@robots.ox.ac.uk\\
\addr University of Oxford}

\begin{document}

\maketitle

\begin{abstract}
Ensembling can improve the performance of Neural Networks, but existing approaches struggle when the architecture likelihood surface has dispersed, narrow peaks.
Furthermore, existing methods construct equally weighted ensembles, and this is likely to be vulnerable to the failure modes of the weaker architectures.
By viewing ensembling as approximately marginalising over architectures we construct ensembles using the tools of Bayesian Quadrature -- tools which are well suited to the exploration of likelihood surfaces with dispersed, narrow peaks.
Additionally, the resulting ensembles consist of architectures weighted commensurate with their performance.
We show empirically -- in terms of test likelihood, accuracy, and expected calibration error -- that our method outperforms state-of-the-art baselines, and verify via ablation studies that its components do so independently.
\end{abstract}

\section{Introduction}
Neural Networks (NNs) are extremely effective function approximators. Their architectures are, however, typically designed by hand, a painstaking process.
Therefore, there has been significant interest in the automatic selection of NN architectures.
In addition to a search strategy, this involves defining a search space from which to select the architecture -- a non-trivial which is also an active area of research.
Recent work shows ensembles of networks of different architectures from a given search space can outperform the \emph{single} best architecture or ensembles of networks of the same architecture \citep{zaidi_neural_2022, shu_neural_2022}. Finding the single best architecture is typically referred to as Neural Architecture Search (NAS) \citep{zoph2016neural, elsken_neural_2019, he2021automl}.
Such ensembles improve performance on a range of metrics, including the test set's predictive accuracy, likelihood, and expected calibration error.
The latter two metrics measure the quality of the model's uncertainty estimates, which have been shown for single architectures to be poor \citep{guo_calibration_2017}.
Performant models in these metrics are crucial for systems which make critical decisions, such as self-driving vehicles.
Ensemble selection is an even more difficult problem to tackle manually than selecting a single architecture, as it requires a combinatorial search over the same space.
Hence, interest in methods for automatic ensemble construction is growing.
This paper targets exactly this problem.

Conceptually, Neural Ensemble Search (NES) algorithms can be split into two stages.
The first is the \emph{candidate selection} stage, which seeks to characterise the posterior distribution, $p(\alpha \mid D)$, given the training data $D$, over architectures from a given search space $\alpha \in \mathcal{A}$.
Multiple approaches have been proposed.
One such is an evolutionary strategy which seeks the modes of this distribution \citep{zaidi_neural_2022}. An alternative is training a ``supernet'' and using it to learn the parameters of a variational approximation to this distribution \citep{shu_neural_2022}.
This involves evaluating the likelihood of a set of architectures from the search space, an evaluation which requires first training the architecture weights.
The second stage is \emph{ensemble selection}, where the ensemble members are selected from the candidate set and each member's weight is chosen.
Several approaches have also been suggested for ensemble selection, such as beam search and sampling from the (approximate) posterior over architectures.

In this work, we investigate novel approaches to both stages of a NES algorithm.
We view ensembling, the averaging over architectures, as marginalisation with respect to a particular distribution over architectures.
When this distribution is the posterior over architectures, we are taking the hierarchical Bayesian approach.
The key advantage of this approach is the principled accounting of uncertainty, which also improves accuracy by preventing overconfidence in a single architecture.
Additionally, this paradigm allows us to bring the tools of Bayesian Quadrature to bear upon the problem of Neural Ensemble Search.
Specifically, the contributions of this work are as follows:\footnote{An implementation of our proposals can be found at \href{https://github.com/saadhamidml/bq-nes}{https://github.com/saadhamidml/bq-nes}.}
\begin{itemize}
\item We propose using an acquisition function for adaptive Bayesian Quadrature to select the candidate set of architectures to train.
It is from this candidate set that the ensemble members are later selected.
\item We show how recombination of the approximate posterior over architectures can be used to construct a weighted ensemble from the candidate set.
\item We undertake an empirical comparison of our proposals against state-of-the-art baselines.
Additionally, we conduct ablation studies to understand the effect of our proposals for each stage of the NES pipeline.
\end{itemize}

\section{Background}
\subsection{Neural Architecture Search (NAS)}
NAS aims to automatically discover high-performing NN architectures and has shown promising performance in various tasks \citep{real2017large, zoph2018learning, liu2019auto}. It is typically formulated as an optimisation problem, i.e. maximising some measure of performance $f$ over a space of NN architectures $\mathcal{A}$,
\begin{equation}
\alpha_* = \text{argmax}_{\alpha \in \mathcal{A}} f(\alpha).
\end{equation}
\citet{elsken_neural_2019} identify three conceptual elements of a NAS pipeline: a search space, a search strategy, and a performance estimation strategy.

The first part of a NAS pipeline -- the search space -- is the way in which the possible space of NN architectures is defined.
In this paper we require the search space be such that a Gaussian Process (GP) can be defined upon it.
In particular, we focus on the two types of search space most common in the literature.
The first is a \textit{cell-based} search space, which consists of architectures made by swapping out ``cells'' in a fixed macro-skeleton \citep{pham2018efficient, dong_nats-bench_2021, liu_darts_2019}.
These cells are represented as directed acyclic graphs where each edge corresponds to an operation from a pre-defined set of operations.
Typically, the macro-skeleton will be structured so that repeating copies of the same cell are stacked within the macro-skeleton.
This structure allows for the representation of the architecture by the corresponding cell.
The second is a \textit{macro} search space defined by varying structural parameters such as kernel size, number of layers, and layer widths.
An example of such a search space is the \textit{Slimmable network} \citep{yu_slimmable_2019, yu2019universally} on the MobileNet search space \citep{sandler2018mobilenetv2}: the largest possible network is trained and all other networks in the search space are given as sub-networks or ``slices''.

A NAS pipeline's second phase is the search strategy.
This is a procedure for selecting which architectures to query the performance of.
All strategies will exhibit an exploration-exploitation trade-off, where exploration is covering the search space well, and exploitation is selecting architectures that are similar to the well-performing architectures in the already queried history.

The final element of a NAS pipeline is the performance estimation strategy, which is the method for querying the performance of a given architecture.
Typically, this is done by training the NN weights, given the architecture, on a training dataset, and evaluating its performance on a validation set.
This demands significant computation and practically limits the total number of architecture evaluations available to the search strategy.
However, for some search spaces -- where network weights are shared -- performance estimation is considerably cheaper.

There exists a large body of literature devoted to NAS, pursuing a range of strategies such as one-shot NAS \citep{liu_darts_2019,xu_pc-darts_2020,chen_progressive_2019, yu2020bignas, bender2018understanding}, evolutionary strategies \cite{real2017large, liang2018evolutionary, liu2021survey}, and Bayesian Optimisation.

\subsection{Bayesian Optimisation for Neural Architecture Search}
An effective approach to NAS is Bayesian Optimisation (BO) \citep{kandasamy_neural_2019,white_bananas_2020,ru_interpretable_2021, Wan_2022_WACV, shi2020bridging, zhou2023autopeft}.
On a high level, BO models the objective function $f$ and sequentially selects where to query next based on an acquisition function, with the goal of finding the optimal value of the objective function in a sample efficient manner.
Typically, the objective function is modelled using a Gaussian Process (GP) -- a stochastic process for which all finite subsets of random variables are joint normally distributed \citep{rasmussen_gaussian_2006}.

A GP is defined using a mean function $m(\alpha)$ that specifies the prior mean at $\alpha$, and a kernel function $k(\alpha, \alpha')$ that specifies the prior covariance between $f(\alpha)$ and $f(\alpha')$.
The posterior, conditioned on a set of observations $A = \{(\alpha_i, )\}_i^N$ and $y = [f(\alpha_1), \ldots, f(\alpha_N)]^T$, is also a GP with moments
\begin{align}
m_A(\cdot) &= m(\cdot) + K_{\cdot A} K_{AA}^{-1} \left( y - m(\alpha)\right) \qquad \text{and} \\
k_A(\cdot, \cdot') &= K_{\cdot \cdot'} - K_{\cdot A} K_{AA}^{-1} K_{A \cdot},
\end{align}
where $K_{XY}$ indicates a matrix generated by evaluating the kernel function between all pairs of points in the sets $X$ and $Y$.
The prior mean function $m$ is typically set to zero.

\citet{ru_interpretable_2021} showed that the Weisfeiler-Lehman graph kernel (WL kernel) \citep{shervashidze_weisfeiler-lehman_2011} is an appropriate choice for modelling NN performance metrics on a cell-based NAS search space with a GP.
To apply the WL kernel, a cell first needs to be represented as a labelled DAG.
Next, a feature vector is built up by aggregating labels for progressively wider neighbourhoods of each node, and building a histogram of the resulting aggregated labels.
The kernel is then computed as the dot product of the feature vectors for a pair of graphs.

A common acquisition function for BO is Expected Improvement \citep{garnett_bayesian_2021},
\begin{equation}
a_{EI}(\alpha) = \mathbb{E}_{p(f \mid D)} \Bigl[\max\bigl(f(\alpha) - f(\hat{\alpha}), 0\bigr)\Bigr]
\end{equation}
where $\hat{\alpha}$ is the best architecture found so far.
Using this acquisition function in conjunction with a GP using the WL kernel was shown by \citet{ru_interpretable_2021} to be effective for NAS.

\subsection{Neural Ensemble Search}
Neural Ensemble Search \citep{zaidi_neural_2022} is a method for automatically constructing ensembles of a given size, $M$, from a NAS search space $\mathcal{A}$.
First, a candidate set of architectures, $A \subset \mathcal{A}$, is selected using a regularised evolutionary strategy (NES-RE), or random sampling from the search space.
The authors propose several ensemble selection methods to subsequently select a subset of $M$ architectures $A_M \subset A$.
Of particular interest in this work are Beam Search (BS) and Weighted Stacking (WS).

BS initially adds the best performing architecture to the ensemble and greedily adds the architecture from the candidate set (without replacement) that most improves the validation loss of the ensemble.
WS optimises the ensemble weights over the whole candidate set on the validation loss (subject to the weights being non-negative and summing to one).
The members with the highest $M$ weights are included in the ensemble, and their corresponding weights renormalised.
The authors compare BS to WS on the CIFAR-10 dataset, and find performance in terms of the log likelihood of the test set to be better for BS for small ensembles, but similar for larger ensembles.

Neural Ensemble Search via Bayesian Sampling \citep{shu_neural_2022} approximates the posterior distribution over architectures $p(\alpha \mid D)$ with a variational distribution of the form $q(\alpha) = \prod_i q_i (o \mid \boldsymbol{\theta}_i)$, where $i$ iterates over the connections within a cell, $o$ is the operation for connection $i$, and $\boldsymbol{\theta}_i$ are the variational parameters for $q_i$.
The form of $q_i$ is chosen to be a softmax over $\boldsymbol{\theta}_i$.
The ensemble is then selected by using Stein Variational Gradient Descent with Regularised Diversity to select a diverse set of $M$ samples from (a continuous relaxation of) the variational distribution.

Relatedly, DeepEnsembles \citep{lakshminarayanan_simple_2017} seeks to approximately marginalise over the parameters of a given NN architecture.
The architecture is trained from several random initialisations, and the ensemble makes a prediction as an equally weighed sum of these.
This is orthogonal to the work above (and, indeed, is orthogonal our work), which seeks to construct ensembles of different architectures, rather than ensembles of different parameter settings of the same architecture.

\subsection{Bayesian Quadrature}
Bayesian Quadrature (BQ) \citep{ohagan_bayeshermite_1991,minka_deriving_2000} is a probabilistic numerical \citep{hennig_probabilistic_2022} integration technique that targets the computation of $Z = \int f(\cdot) d\pi(\cdot)$ based on evaluations of the integrand $f$ (assuming a given prior $\pi$).
Similar to BO, it maintains a surrogate model over the integrand $f$, which induces a posterior over the integral value $Z$.
BQ also makes use of an acquisition function to iteratively select where next to query the integrand.

The surrogate model for BQ is usually chosen to be a GP, and this induces a Gaussian posterior over $Z \sim \mathcal{N} (\mu_Z, \sigma_Z)$.
The moments of this posterior are given by
\begin{align}
\mu_Z \!&=\! \int K(\cdot, X) d\pi(\cdot) K_{XX}^{-1} f, \quad \text{and} \\
\sigma_Z \!&=\! \int K(\cdot, \cdot') - K(\cdot, X) K_{XX}^{-1} K(X, \cdot) d\pi(\cdot) d\pi(\cdot'),
\end{align}
where $X$ is the set of query points, and $f$ are the corresponding integrand observations.
Note that the posterior mean $\mu_Z$ takes the form of a quadrature rule -- a weighted sum of function evaluations $\sum_i w_i f(x_i)$ where $w_i$ are the elements of the vector $\int K(\cdot, X) d\pi(\cdot) K_{XX}^{-1}$.

Frequently, the integrand of interest is non-negative.
Important examples of such integrands are likelihood functions (which are integrated with respect to a prior to compute a model evidence) and predictive densities (which are integrated with respect to a posterior to compute a posterior predictive density).
Warped Bayesian Quadrature \citep{osborne_active_2012,gunter_sampling_2014,chai_improving_2019} allows practitioners to incorporate the prior knowledge that the integrand is non-negative into the surrogate model.
Of particular interest in this work will be the WSABI-L model \citep{gunter_sampling_2014}, which models the square-root of the integrand with a GP, $\sqrt{2\bigl(f(x) - \beta\bigr)} \sim \mathcal{GP} \bigl(\mu_D(x), \Sigma_D(x, x')\bigr)$.
This induces a (non-central) chi-squared distribution over $f$ which can be approximated with a GP, with moments
\begin{align}
m(x) &= \beta + \frac{1}{2} \mu_D(x)^2, \\
k(x, x') &= \mu_D(x) \Sigma_D(x, x') \mu_D(x').
\end{align}

\citet{gunter_sampling_2014} established, empirically, that the uncertainty sampling acquisition function works well for Bayesian Quadrature.
This acquisition function targets the variance of the integrand
\begin{align}
a_{US}(x) = \Sigma_D(x, x) \mu_D(x)^2 \pi(x)^2.
\end{align}
This naturally trades off between exploration (regions where $\Sigma_D(x, x)$ is high), and exploitation (regions where $\mu_D(x)$ is high -- most of the volume under the integrand is concentrated here).

Just as BO is a natural choice for NAS -- an expensive black-box optimisation problem -- so BQ is a natural choice for NES -- an expensive black-box marginalisation problem.
It is this realisation that inspires our proposals in Section \ref{sec:method}.

\subsection{Recombination}
\label{sec:recombination}
Given a non-negative measure supported on $N$ points $\{(w_n, x_n)\}_{n=1}^N$ where $w_n \geq 0$ and $\sum_{n=1}^N w_n = 1$, and $M - 1$ ``test'' functions $\{\phi_t(\cdot)\}_{t=1}^{M-1}$, it is possible to find a subset of $M < N$ points $\{x_n\}_{m=1}^M \subset \{x_n\}_{n=i}^N$ for which
\begin{equation}
\sum_{m=1}^M w_m \phi_t(x_m) = \sum_{n=1}^N w_n \phi_t(x_n)
\end{equation}
for all $\phi_t$, with $w_m \geq 0$ and $\sum_{m=1}^M w_m = 1$ \citep{tchernychova_caratheodory_2015}.

For Kernel Quadrature, one can use the Nystr\"om approximation of the kernel matrix to obtain a set of test functions \citep{hayakawa_positively_2022,adachi_fast_2022}.
Using a subset, $S$, of $M - 1$ data points, the kernel can be approximated $\tilde{k}(x, x') = k(x, S) k(S, S)^{-1} k(S, x')$.
By taking an eigendecomposition, $k(S, S) = U \Lambda U^T$, the approximate kernel can be expressed as
\begin{equation}
\label{eq:nystrom_test_functions}
\tilde{k}(x, x') = \sum_t^{M - 1} \frac{1}{\lambda_t} \bigl(u_t^T k(S, x)\bigr) \bigl(u_t^T k(S, x')\bigr)
\end{equation}
where $u_i$ are the columns of $U$, and $\lambda_i$ the diagonal elements of $\Lambda$.
We can then use $\phi_t(\cdot) = u_t^T k(S, \cdot)$ as test functions.

\section{Bayesian Quadrature for Neural Ensemble Search}
\label{sec:method}

\begin{figure*}[t]
\centering
\includegraphics[width=\hsize]{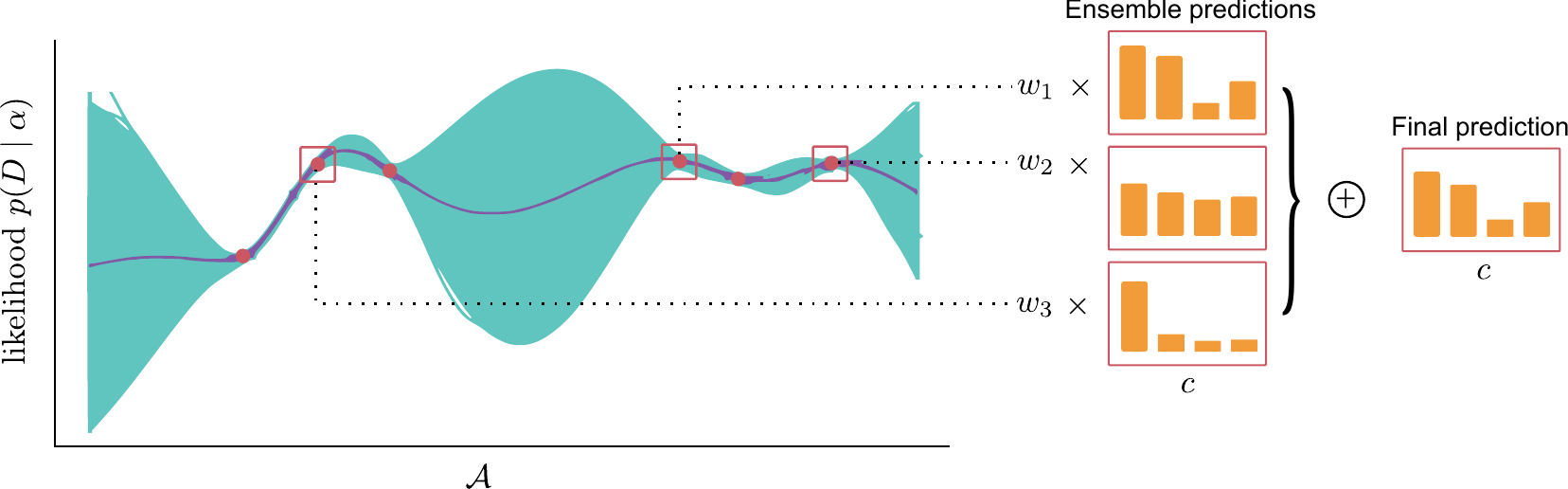}
\caption{A schematic representation of our proposal.
The plot on the left shows a Gaussian Process modelling the likelihood over the space of architectures.
The architectures to train and evaluate the likelihood for are selected by maximising a Bayesian Quadrature acquistion function, as described in Section \ref{sec:candidate_selection}. 
One of the algorithms described in Section \ref{sec:ensemble_selection} is then used to select the subset of architectures to include in the ensemble, along with their weights.
The final prediction is then a linear combination of the predictions of each ensemble member, as shown by the bar plots on the right (where each bar indicates the probability assigned to a particular class).}
\label{fig:infographic}
\end{figure*}

We decompose NES into two sub-problems:
\begin{enumerate}
\item The selection of a candidate set of architectures $\{\alpha_i\}_{i=1}^N = A \subset \mathcal{A}$ for which to train the architecture parameters.
\item The selection of a set of $M$ members from the candidate set to include in the ensemble, and their weights, $\boldsymbol{w} \in \mathbb{R}^M$.
\end{enumerate}
We take novel approaches to each of these sub-problems, described respectively in the following two subsections.
Algorithms \ref{alg:candidate_selection}, \ref{alg:ensemble_selection_bq} and \ref{alg:ensemble_selection_rs} summarise our propositions.

\subsection{Building the Candidate Set}
\label{sec:candidate_selection}
An ensemble's prediction is a weighted sum of the predictions of its constituent members, $A_M$, and this can always be viewed as approximating an expectation with respect to a distribution, $\pi$, over architectures,
\begin{equation}
\label{eq:ensemble_marginalise}
\mathbb{E}_{\pi(\alpha)} \bigl[ p(c \mid x, \alpha) \bigr]= \sum_{\alpha \in \mathcal{A}} p(c \mid x, \alpha) \pi(\alpha) \approx \sum_{\alpha \in A_M} p(c \mid x, \alpha) \pi_M(\alpha).
\end{equation}
$p(c \mid x, \alpha, D)$ is the predictive probability assigned to class $c \in \{1, \ldots, C\}$ by the architecture $\alpha$ (conditioned on the training data $D$) given the input $x \in \mathcal{X}$.
The expectation in Equation \eqref{eq:ensemble_marginalise} corresponds to a marginalisation over architectures.
The set of architectures $A_M$ and their weights $\pi_M$ can be seen as a quadrature rule to approximate this marginalisation.
When $\pi$ is the posterior over architectures $p(\alpha \mid D)$, we are performing hierarchical Bayesian inference, and the result is the posterior predictive distribution,
\begin{align}
p(c \mid x, D) &= \sum_{\alpha \in \mathcal{A}} p(c \mid x, \alpha, D) \, p(\alpha \mid D) \nonumber \\
&= \frac{\sum_{\alpha \in \mathcal{A}} p(c \mid x, \alpha, D) \, p(D \mid \alpha) \, p(\alpha)}{\sum_{\alpha \in \mathcal{A}} p(D \mid \alpha) \, p(\alpha)}. \label{eq:posterior_predictive}
\end{align}

By taking the view of ensembling as marginalisation, the practitioner has the ability to make their belief over $\mathcal{A}$ explicit.
As the training data is finite, it is rarely appropriate to concentrate all of the probability mass of $\pi$ on a single architecture.
Arguably, $\pi(\alpha) = p(\alpha \mid D)$ is the most appropriate choice of $\pi$ as it is the distribution implied by the prior and the architectures' ability to explain the observed data.
The hierarchical Bayesian framework should offer the most principled accounting of uncertainty in the choice of architecture (a more concentrated $\pi$ should over-fit, a less concentrated $\pi$ should under-fit).

From Equation \eqref{eq:posterior_predictive} we see that, to compute the posterior predictive, we need to compute $C$ sums of products of functions of the architecture and the architecture likelihoods.
Intuitively, we expect a quadrature scheme that approximates well the sum in the denominator of \eqref{eq:posterior_predictive} will also approximate the sum in the numerator well.
Therefore, we propose using a Bayesian Quadrature acquisition function to build up the candidate set, as these architectures will form the nodes of a query-efficient quadrature scheme for \eqref{eq:posterior_predictive} and so a good basis for an ensemble.

The likelihood of an architecture $p(D \mid \alpha)$ is not typically available, as this would require marginalisation over the NN weights, $w$, of the architecture.
We instead approximate using the MLE, which is equivalent to assuming the prior of the architecture weights is a Dirac delta delta distribution at the maximiser of the (architecture weights') likelihood function.
\begin{align}
p(D \mid \alpha) &= \int p(D \mid w, \alpha) \, p(w \mid \alpha) dw 
\approx p(D \mid \hat{w}, \alpha), \\
\hat{w} &= \text{argmax}_w p(D \mid w, \alpha) p(w \mid \alpha).
\end{align}
Computing $p(c \mid x, \alpha, D)$ requires an analogous intractable marginalisation.
We approximate it similarly, noting that it depends only indirectly on the training data, through the optimisation procedure, i.e.
\begin{align}
p(c \mid x, \alpha, D) &= \int p(c \mid x, w, \alpha, D) \, p(w \mid \alpha, D) dw \nonumber\\
&\approx p(c \mid x, \hat{w}, \alpha).
\end{align}

Concretely, we place a functional prior on the architecture likelihood surface, warped using the square-root transform, $\sqrt{2 \bigl( p(D \mid \hat{w}, \alpha) - \beta \big)} \sim \mathcal{GP}$, and use uncertainty sampling to make observations of the likelihood at a set of architectures $\{\alpha_i\}_{i=1}^N = A \subset \mathcal{A}$.

This provides us with an estimate of the model evidence $Z = \sum_{\alpha \in \mathcal{A}} p(D \mid \hat{w}, \alpha) \, p(\alpha)$, which we denote $\hat{Z}$.
The computation of this estimate requires Monte Carlo sampling to approximate sums of (products of) the WL-kernel over $\mathcal{A}$.
Note this is far more feasible than approximating the original sums in Equation \eqref{eq:posterior_predictive} with Monte Carlo sampling as $K(\alpha_j, A)$ is far cheaper to evaluate than $p(D \mid \hat{w}, \alpha_j)$ or $p(c \mid x, \hat{w}, \alpha_j)$ -- either would require training architecture $\alpha_j$.

\begin{algorithm}
\caption{Candidate set selection algorithm using a BQ acquisition function.
Returns architectures $A = \{\alpha_i\}_{i=1}^N$ and their corresponding likelihoods $L = \{p(D \mid \hat{w}, \alpha_i)\}_{i=1}^N$.}
\label{alg:candidate_selection}
\begin{algorithmic}
\State $A, L \gets \text{sample}(n, \mathcal{A})$ \Comment{Initial samples.}
\State $\theta \gets \text{argmax}_{\theta} p(L \mid A, \theta)$ \Comment{Optimise WL kernel.}
\While{$i > 0$}
\State $\alpha \gets \text{argmax}_{\alpha \in \mathcal{A}} \text{acquisition\_function}(\alpha, A, L, \theta)$
\State $A \gets \{A, \alpha\}$
\State $L \gets \{L, p(D \mid \hat{w}, \alpha)\}$
\State $\theta \gets \text{argmax}_{\theta} p(L \mid A, \theta)$
\EndWhile\\
\Return $A, L$
\end{algorithmic}
\end{algorithm}

\subsection{Selecting the Ensemble}
\label{sec:ensemble_selection}
In principle, the ensemble can be constructed using the weights provided by the quadrature scheme, as these weights naturally trade-off between member diversity and member performance.
However, we wish to select a subset of the candidate set for the ensemble (as it is assumed that an ensemble of the whole candidate set is too costly to be practical for deployment).
Concretely, we seek a subset $A_M \subset A$, along with weights $\boldsymbol{w} \in \mathbb{R}^M$ such that
\begin{align}
p(c \mid x, D) &\approx \sum_n^N \frac{1}{\hat{Z}} \, p(D \mid \alpha_n) \, p(\alpha_n) \, p(c \mid x, \alpha_n, D) + \epsilon \\
&\approx \sum_m^M w_m p(c \mid x, \alpha_m, D) + \epsilon.
\end{align}
We expect $\epsilon$ to be small if regions of high likelihood have been well-explored by the acquisition function in the building of the candidate set.
To select the weights $\boldsymbol{w}$ and the set $A_M$ we can use any recombination algorithm, using the Nystr\"om approximation to generate the test functions, as described in Section \ref{sec:recombination}, and the estimated posterior over architectures as the measure to recombine.
We refer to this algorithm as Posterior Recombination (PR).

A second approach, which we refer to as Re-weighted Stacking (RS), is a modification of Weighted Stacking.
Like for WS, we optimise the weights of an ensemble of the whole candidate set to minimize the validation loss.
The ensemble members are then chosen by selecting the members with the $M$ highest weights.
However, rather than renormalising the corresponding weights, as suggested in \citet{zaidi_neural_2022}, we reallocate the weight assigned to excluded architectures proportionally to the relative covariance between them and the ensemble members.
Concretely, let $\{(\alpha_m, \omega_m)\}_{m=1}^M$ be the ensemble members and their optimised weights, and $\{(\alpha_l, \omega_l)\}_{l=1}^{N-M}$ be the excluded architectures and their optimised weights.
The weights of the ensemble $\boldsymbol{w} \in \mathbb{R}^M$ are given by
\begin{equation}
\label{eq:reweight}
\boldsymbol{w}_m = \omega_m + \sum_{l=1}^{N-M} \frac{k(\alpha_m, \alpha_l)}{\sum_{m=1}^M k(\alpha_m, \alpha_l)} \omega_l.
\end{equation}
\begin{algorithm}
\caption{Posterior recombination.}
\label{alg:ensemble_selection_bq}
\begin{algorithmic}
\State $T \gets \text{nystrom\_test\_functions}(K_{AA}, A)$ \Comment{From Eq \eqref{eq:nystrom_test_functions}}
\State $\mu \gets \left[\frac{p(D \mid \alpha_n) p(\alpha_n)}{\hat{Z}}\right]_{n=1}^N$
\State $\boldsymbol{w}, A_M \gets \text{recombination}(T, \mu)$
\end{algorithmic}
\end{algorithm}

\begin{algorithm}
\caption{Re-weighted stacking.}
\label{alg:ensemble_selection_rs}
\begin{algorithmic}
\State $\omega \gets \text{argmin}_{\omega \in \Delta} \text{loss}(\sum_i \omega_i p(c \mid x, \alpha_n, D), D_{\text{val}})$
\State $I \gets \text{select\_top}(M, \omega)$ \Comment{Select top M.}
\For{$m$ in I}
\State $\boldsymbol{w}_m \gets \text{reweight}(m, I, \omega, k(A, A))$ \Comment{Eq \eqref{eq:reweight}.}
\EndFor
\end{algorithmic}
\end{algorithm}

Our proposals can be combined to yield two possible algorithms.
Both share the same candidate selection strategy that uses a WSABI-L surrogate model with the uncertainty sampling acquisition function to select the set of architectures to train (Algorithm \ref{alg:candidate_selection}).
``BQ-R'' then uses posterior recombination (Algorithm \ref{alg:ensemble_selection_bq}) to select a subset of architectures from the candidate set to include in the ensemble, and choose their corresponding weights.
``BQ-S'' instead uses re-weighted stacking (Algorithm \ref{alg:ensemble_selection_rs} to select, and weight, the ensemble members from the candidate set.
Note that ``BQ-R'' performs approximate hierarchical Bayesian inference using BQ, but ``BQ-S'' is a heuristic inspired by BQ.
Figure \ref{fig:infographic} is a schematic representation of these algorithms.

\section{Experiments}
We begin by performing comparisons on the NATS-Bench benchmark \citep{dong_nats-bench_2021}.
Specifically, we use the provided topology search space, which consists of cells with 4 nodes, 6 connections, and 5 possible operations (including ``zeroise'' which is equivalent to removing a connection) in a fixed macro-skeleton.
The architecture weights are trained for 200 epochs on the CIFAR-100 and ImageNet16-120 (a smaller version of ImageNet with $16 \times 16$ pixel input images, and 120 classes) datasets.
We will compare ensemble performance as measured by test accuracy, test likelihood, and expected calibration error on the test set for a range of ensemble sizes.

First we verify that our chosen surrogate model (WSABI-L) performs well.
Table \ref{tab:gp_posteriors} shows model performance, measured by root mean square error (RMSE) and negative log predictive density (NLPD) on a test set.
The test set is selected by ranking all the architectures in the search space by validation loss, and selecting every 25th architecture.
This ensures that the test set contains architectures across the full range of performance.
We build on the results of \citet{ru_interpretable_2021}, who showed that a GP with a WL kernel is able to model the architecture likelihood surface well.
Our results show that WSABI-L (with a WL kernel) is a consistently better model than an ordinary GP (with a WL kernel).

\begin{table}[t]
\centering
\resizebox{0.8\linewidth}{!}{%
\begin{tabular}{lcccc}
\toprule
& \multicolumn{2}{c}{CIFAR-100} & \multicolumn{2}{c}{ImageNet16-120} \\
\cmidrule(lr){2-3} \cmidrule(lr){4-5}
Model & RMSE & NLPD & RMSE & NLPD \\
\midrule
GP & 6.165 $\pm$ 0.116 & 0.124 $\pm$ 0.013 & 9.610 $\pm$ 0.626 & 0.121 $\pm$ 0.012 \\
WSABI-L & \textbf{5.797 $\pm$ 0.043} & \textbf{-2.741 $\pm$ 0.095} & \textbf{4.078 $\pm$ 0.058} & \textbf{-3.437 $\pm$ 0.040} \\
\bottomrule
\end{tabular}
}
\caption{The (normalised) RMSE and NLPD of a WSABI-L surrogate and a GP surrogate on the test sets.}
\label{tab:gp_posteriors}
\end{table}

Next, we examine the effect of the candidate selection algorithm, shown in Table \ref{tab:candidate_comparison}.
In all cases, we use our variant of weighted stacking, described in Section \ref{sec:ensemble_selection}, to select and weight the ensemble members.
We compare Expected Improvement (EI) with a GP surrogate with a WL kernel, Uncertainty Sampling with a WSABI-L surrogate using a WL kernel (US), and Regularised Evolution (RE).
We find that the US candidate set performs best for ImageNet16-120 in terms of accuracy and LL, but that the RE candidate set performs best for ECE on ImageNet16-120, and across all metrics for CIFAR-100.

\begin{table*}[t]
\centering
\resizebox{0.95\linewidth}{!}{%
\begin{tabular}{lcccccc}
\toprule
& \multicolumn{3}{c}{CIFAR-100} & \multicolumn{3}{c}{ImageNet16-120} \\
\cmidrule(lr){2-4} \cmidrule(lr){5-7}
Algorithm & Accuracy & ECE & LL & Accuracy & ECE & LL \\
\midrule
\textbf{$M = 3$} \\
RE & \textbf{77.1 $\pm$ 0.2} & \textbf{0.018 $\pm$ 0.001} & \textbf{-4385 $\pm$ 24.89} & 51.9 $\pm$ 0.2 & \textbf{0.029 $\pm$ 0.002} & -5595 $\pm$ 12.15 \\
EI & 76.1 $\pm$ 0.2 & 0.024 $\pm$ 0.001 & -4472 $\pm$ 29.74 & 51.4 $\pm$ 0.2 & 0.034 $\pm$ 0.002 & -5632 $\pm$ 11.91 \\
US & 76.6 $\pm$ 0.2 & 0.021 $\pm$ 0.001 & \textbf{-4417 $\pm$ 35.85} & \textbf{52.2 $\pm$ 0.1} & \textbf{0.029 $\pm$ 0.001} & \textbf{-5543 $\pm$ 10.87} \\
\midrule
\textbf{$M = 5$} \\
RE & \textbf{78.5 $\pm$ 0.2} & \textbf{0.033 $\pm$ 0.001} & \textbf{-4013 $\pm$ 19.08} & 53.3 $\pm$ 0.2 & \textbf{0.043 $\pm$ 0.002} & -5417 $\pm$ 12.90 \\
EI & 77.4 $\pm$ 0.2 & 0.039 $\pm$ 0.001 & -4126 $\pm$ 22.25 & 52.6 $\pm$ 0.3 & 0.053 $\pm$ 0.003 & -5479 $\pm$ 15.55 \\
US & 77.8 $\pm$ 0.2 & 0.040 $\pm$ 0.002 & -4077 $\pm$ 33.60 & \textbf{53.6 $\pm$ 0.1} & 0.050 $\pm$ 0.002 & \textbf{-5380 $\pm$ 12.31} \\
\midrule
\textbf{$M = 10$} \\
RE & \textbf{79.4 $\pm$ 0.1} & \textbf{0.053 $\pm$ 0.002} & \textbf{-3759 $\pm$ 16.38} & \textbf{54.5 $\pm$ 0.2} & \textbf{0.065 $\pm$ 0.002} & \textbf{-5280 $\pm$ 16.85} \\
EI & 78.2 $\pm$ 0.2 & \textbf{0.055 $\pm$ 0.002} & -3889 $\pm$ 23.79 & 53.4 $\pm$ 0.2 & 0.071 $\pm$ 0.002 & -5368 $\pm$ 19.47 \\
US & 78.6 $\pm$ 0.2 & 0.059 $\pm$ 0.001 & -3843 $\pm$ 22.71 & \textbf{54.7 $\pm$ 0.1} & 0.072 $\pm$ 0.001 & \textbf{-5262 $\pm$ 9.964} \\
\bottomrule
\end{tabular}
}
\caption{Test accuracy, expected calibration error, and log likelihood on CIFAR-100 and ImageNet16-120 for our candidate set selection method (US) and baselines.
The numbers shown are means and standard error of the mean over 10 repeats.
Each candidate set selection method is initialised with 10 random architectures, and used to build a set of 150 architectures.
The ensemble is chosen and weighted using our variant of weighted stacking.
We see that the RE candidate set performs best for CIFAR-100, and in terms of ECE for ImageNet16-120.
The US candidate set performs best in terms of accuracy and LL for ImageNet16-120.}
\label{tab:candidate_comparison}
\end{table*}

We then move on to comparing the effect of the ensemble selection algorithm, shown in Table \ref{tab:ensemble_comparison}.
In all cases, we use uncertainty sampling with a WSABI-L surrogate to build the candidate set.
We initialise with 10 architectures randomly selected from a uniform prior over the search space, and use the acquisition function to build a set of 150 architectures.
We compare beam search (BS), weighted stacking (WS), recombination of the approximate posterior (PR), and re-weighted stacking (RS).
We find that the stacking variants consistently perform best (with RS slightly improving upon WS) in terms of accuracy and LL, and PR in terms of ECE for larger datasets.

\begin{table*}[t]
\vspace{-4mm}
\centering
\resizebox{0.95\linewidth}{!}{%
\begin{tabular}{lcccccc}
\toprule
& \multicolumn{3}{c}{CIFAR-100} & \multicolumn{3}{c}{ImageNet16-120} \\
\cmidrule(lr){2-4} \cmidrule(lr){5-7}
Algorithm & Accuracy & ECE & LL & Accuracy & ECE & LL \\
\midrule
\textbf{$M = 3$} \\
BS & 75.2 $\pm$ 0.2 & 0.030 $\pm$ 0.002 & -4500 $\pm$ 41.04 & \textbf{52.2 $\pm$ 0.1} & 0.036 $\pm$ 0.002 & -5572 $\pm$ 13.17 \\
WS & \textbf{76.4 $\pm$ 0.2} & \textbf{0.021 $\pm$ 0.001} & \textbf{-4426 $\pm$ 35.87} & \textbf{52.1 $\pm$ 0.1} & \textbf{0.029 $\pm$ 0.001} & \textbf{-5545 $\pm$ 10.57} \\
PR & 71.9 $\pm$ 0.8 & 0.075 $\pm$ 0.025 & -5259 $\pm$ 300.9 & 46.7 $\pm$ 2.3 & 0.052 $\pm$ 0.021 & -6347 $\pm$ 480.5 \\
RS & \textbf{76.6 $\pm$ 0.2} & \textbf{0.021 $\pm$ 0.001} & \textbf{-4417 $\pm$ 35.85} & \textbf{52.2 $\pm$ 0.1} & \textbf{0.029 $\pm$ 0.001} & \textbf{-5543 $\pm$ 10.87} \\
\midrule
\textbf{$M = 5$} \\
BS & 76.4 $\pm$ 0.2 & 0.048 $\pm$ 0.002 & -4233 $\pm$ 36.48 & \textbf{53.4 $\pm$ 0.1} & 0.058 $\pm$ 0.002 & -5410 $\pm$ 10.70 \\
WS & \textbf{77.7 $\pm$ 0.2} & \textbf{0.036 $\pm$ 0.002} & \textbf{-4088 $\pm$ 34.13} & \textbf{53.6 $\pm$ 0.1} & 0.049 $\pm$ 0.001 & \textbf{-5382 $\pm$ 12.55} \\
PR & 73.3 $\pm$ 0.9 & \textbf{0.040 $\pm$ 0.004} & -4768 $\pm$ 174.3 & 50.7 $\pm$ 0.3 & \textbf{0.028 $\pm$ 0.004} & -5647 $\pm$ 50.58 \\
RS & \textbf{77.8 $\pm$ 0.2} & \textbf{0.040 $\pm$ 0.002} & \textbf{-4077 $\pm$ 33.60} & \textbf{53.6 $\pm$ 0.1} & 0.050 $\pm$ 0.002 & \textbf{-5380 $\pm$ 12.31} \\
\midrule
\textbf{$M = 10$} \\
BS & 76.9 $\pm$ 0.3 & 0.063 $\pm$ 0.001 & -4079 $\pm$ 50.29 & 54.1 $\pm$ 0.1 & 0.076 $\pm$ 0.001 & -5307 $\pm$ 9.795 \\
WS & \textbf{78.5 $\pm$ 0.2} & 0.055 $\pm$ 0.002 & \textbf{-3848 $\pm$ 23.96} & \textbf{54.6 $\pm$ 0.1} & 0.070 $\pm$ 0.001 & \textbf{-5264 $\pm$ 10.11} \\
PR & 75.5 $\pm﻿$ 0.9 & \textbf{0.037 $\pm$ 0.002} & -4309 $\pm$ 172.6 & 52.3 $\pm$ 0.3 & \textbf{0.018 $\pm$ 0.001} & -5412 $\pm$ 22.96 \\
RS & \textbf{78.6 $\pm$ 0.2} & 0.059 $\pm$ 0.001 & \textbf{-3843 $\pm$ 22.71} & \textbf{54.7 $\pm$ 0.1} & 0.072 $\pm$ 0.001 & \textbf{-5262 $\pm$ 9.964} \\

\bottomrule
\end{tabular}
}
\caption{Test accuracy, expected calibration error, and log likelihood on CIFAR-100 and ImageNet16-120 for Beam Search (BS), Weighted Stacking (WS), Posterior Recombination (PR), and Re-weighted Stacking (RS).
The numbers shows are means and standard error of the mean over 10 repeats.
The candidate set selection method is our method -- Uncertainty Sampling with a WSABI-L surrogate -- initialised with 10 random architectures, and used to build a set of 150 architectures.
We see that the stacking variants consistently perform best for accuracy and LL, with RS slightly improving upon WS.
For ECE, RS and WS perform well for small ensembles, but PR works best for larger ensembles.}
\label{tab:ensemble_comparison}
\end{table*}

We then proceed to compare the two variants of our alorithm -- BQ-R and BQ-S -- with one simple and two state-of-the-art baselines.
\begin{description}
\item[Random] The ensemble is an evenly weighted combination of $M$ architectures randomly sampled from the prior $p(\alpha)$ over the search space.
\item[NES-RE] The candidate set is selected using regularised evolution, and the ensemble members are chosen using beam search.
The ensemble members are equally weighted.
\item[NES-BS] The posterior over architectures $p(\alpha \mid D)$ is approximated using a variational distribution.
The ensemble is constructed by sampling $M$ architectures from the variational distribution using Stein-Variational Gradient Descent.
As no implementation is publicly available, we provide our own. However, our implementation learns the variational parameters by approximating the expected log likelihood term of the ELBO using a subset of the search space, rather than by backpropagating through a ``supernet'' as described by \cite{shu_neural_2022}. The subset we use are the 150 architectures in the search space with the highest likelihoods on the validation set. (Of course this is only possible when working with a NAS benchmark.) We argue that our approximation is suitable as most posterior mass will be concentrated on these architectures, so a good variational distribution will concentrate mass on them as well. Additionally, our approximation is much faster as it does not require training a supernet.
\end{description}
Table \ref{tab:joint_comparison} presents the results on CIFAR-100 and ImageNet16-120 for a range of ensemble sizes.
Whilst NES-RE matches or does slightly better than our proposals in terms of accuracy and LL on CIFAR-100, we find that both BQ-S and BQ-R often perform better in terms of expected calibration error.
BQ-S achieves the best performance on ImageNet16-120 in terms of LL across all ensemble sizes, is joint best with NES-RE in terms of accuracy, and often outperforms NES-RE in terms of ECE.

\begin{table*}[t]
\centering
\resizebox{0.95\linewidth}{!}{%
\begin{tabular}{lcccccc}
\toprule
& \multicolumn{3}{c}{CIFAR-100} & \multicolumn{3}{c}{ImageNet16-120} \\
\cmidrule(lr){2-4} \cmidrule(lr){5-7}
Algorithm & Accuracy & ECE & LL & Accuracy & ECE & LL \\
\midrule
Best Single & 69.1 & 0.088 & -5871 & 45.9 & 0.062 & -6386 \\
\midrule
\textbf{$M = 3$} \\
Random & 69.2 $\pm$ 1.5 & 0.075 $\pm$ 0.007 & -5778 $\pm$ 291.3 & 39.7 $\pm$ 2.2 & 0.097 $\pm$ 0.007 & -7459 $\pm$ 309.6 \\
NES-RE & \textbf{76.6 $\pm$ 0.2} & 0.026 $\pm$ 0.002 & \textbf{-4340 $\pm$ 19.58} & \textbf{52.0 $\pm$ 0.2} & 0.033 $\pm$ 0.002 & -5582 $\pm$ 8.858 \\
NES-BS & 66.2 $\pm$ 1.5 & 0.073 $\pm$ 0.009 & -6477 $\pm$ 203.0 & 45.7 $\pm$ 0.3 & 0.058 $\pm$ 0.003 & -6403 $\pm$ 28.04 \\
BQ-R & 71.9 $\pm$ 0.8 & 0.075 $\pm$ 0.025 & -5259 $\pm$ 300.9 & 46.7 $\pm$ 2.3 & 0.052 $\pm$ 0.021 & -6347 $\pm$ 480.5 \\
BQ-S & \textbf{76.6 $\pm$ 0.2} & \textbf{0.021 $\pm$ 0.001} & -4417 $\pm$ 35.85 & \textbf{52.2 $\pm$ 0.1} & \textbf{0.029 $\pm$ 0.001} & \textbf{-5543 $\pm$ 10.87} \\
\midrule
\textbf{$M = 5$} \\
Random & 72.2 $\pm$ 0.9 & 0.111 $\pm$ 0.009 & -5304 $\pm$ 180.9 & 42.7 $\pm$ 1.5 & 0.129 $\pm$ 0.008 & -7135 $\pm$ 216.1 \\
NES-RE & \textbf{78.2 $\pm$ 0.1} & \textbf{0.042 $\pm$ 0.002} & \textbf{-4002 $\pm$ 17.11} & \textbf{53.4 $\pm$ 0.2} & 0.051 $\pm$ 0.001 & -5404 $\pm$ 12.59 \\
NES-BS & 65.9 $\pm$ 1.5 & 0.073 $\pm$ 0.009 & -6481 $\pm$ 208.7 & 45.7 $\pm$ 0.3 & 0.058 $\pm$ 0.003 & -6403 $\pm$ 28.04 \\
BQ-R & 73.3 $\pm$ 0.9 & \textbf{0.040 $\pm$ 0.004} & -4768 $\pm$ 174.3 & 50.7 $\pm$ 0.3 & \textbf{0.028 $\pm$ 0.004} & -5647 $\pm$ 50.58 \\
BQ-S & 77.8 $\pm$ 0.2 & \textbf{0.040 $\pm$ 0.002} & -4077 $\pm$ 33.60 & \textbf{53.6 $\pm$ 0.1} & 0.050 $\pm$ 0.002 & \textbf{-5380 $\pm$ 12.31} \\
\midrule
\textbf{$M = 10$} \\
Random & 74.7 $\pm$ 0.3 & 0.150 $\pm$ 0.010 & -5018 $\pm$ 82.21 & 45.1 $\pm$ 0.4 & 0.159 $\pm$ 0.008 & -6916 $\pm$ 73.21 \\
NES-RE & \textbf{79.4 $\pm$ 0.1} & 0.060 $\pm$ 0.001 & \textbf{-3763 $\pm$ 15.16} & \textbf{54.5 $\pm$ 0.2} & 0.069 $\pm$ 0.001 & \textbf{-5269 $\pm$ 17.83} \\
NES-BS & 69.1 $\pm$ 0.4 & 0.085 $\pm$ 0.005 & -6119 $\pm$ 36.31 & 45.6 $\pm$ 0.3 & 0.068 $\pm$ 0.004 & -6442 $\pm$ 24.47 \\
BQ-R & 75.5 $\pm﻿$ 0.9 & \textbf{0.037 $\pm$ 0.002} & -4309 $\pm$ 172.6 & 52.3 $\pm$ 0.3 & \textbf{0.018 $\pm$ 0.001} & -5412 $\pm$ 22.96 \\
BQ-S & 78.6 $\pm$ 0.2 & 0.059 $\pm$ 0.001 & -3843 $\pm$ 22.71 & \textbf{54.7 $\pm$ 0.1} & 0.072 $\pm$ 0.001 & \textbf{-5262 $\pm$ 9.964} \\
\bottomrule
\end{tabular}
}
\caption{Test accuracy, expected calibration error (ECE), and log likelihood (LL) on CIFAR-100 and ImageNet16-120 for our proposals (BQ-R and BQ-S) and baselines.
For reference we also include the performance of the best architecture (measured by validation loss) on the test set (labelled Best Single).
The numbers show are means and standard error of the mean over 10 repeats.
Where applicable, the candidate set selection method is initialised with 10 random architectures, and used to build a set of 150 architectures.
For ImageNet16-120 we see that BQ-S performs best across ensemble sizes in terms of LL, and joint best with NES-RE in terms of accuracy.
For CIFAR-100 we find that NES-RE performs best in terms of accuracy and LL.
Particularly for larger ensembles, BQ-R performs best in terms of ECE.}
\label{tab:joint_comparison}
\end{table*}


Next, we perform a study on a larger search space defined by a ``slimmable network'' \citep{yu_slimmable_2019}, consisting of 614,625 architectures.
Sub-networks or ``slices'' of this supernet constitute architectures within this search space.
The architectures are structured as a chain of 7 blocks, each of which can have up to 4 layers.
These sub-networks can be represented in a 28 dimensional ordinal space (with 4 options along each dimension).
We compare the best performing variant of our method, BQ-S, and the best performing baseline, NES-RE, from the smaller NATS-Bench search space.
We use an RBF kernel with WSABI-L for Uncertainty Sampling with our method BQ-S, and compare to NES-RE.
The results are shown in Table \ref{tab:supernet}.
We see that BQ-S consistently outperforms NES-RE in terms of log likelihood of the test set and, for CIFAR-100, in terms of expected calibration error as well.

\begin{table}[t]
\centering
\resizebox{0.95\linewidth}{!}{%
\begin{tabular}{lcccccc}
\toprule
& \multicolumn{3}{c}{CIFAR-10} & \multicolumn{3}{c}{CIFAR-100} \\
\cmidrule(lr){2-4} \cmidrule(lr){5-7}
Algorithm & Accuracy & ECE & LL & Accuracy & ECE & LL \\
\midrule
\textbf{$M = 3$} \\
NES-RE & 93.8 $\pm$ 0.0 & 0.029 $\pm$ 0.001 & -1165 $\pm$ 5.602 & 74.2 $\pm$ 0.2 & 0.072 $\pm$ 0.004 & -5136 $\pm$ 61.49 \\
BQ-S & 93.7 $\pm$ 0.1 & 0.030 $\pm$ 0.000 & \textbf{-1152 $\pm$ 5.215} & 74.4 $\pm$ 0.1 & \textbf{0.063 $\pm$ 0.002} & \textbf{-5021 $\pm$ 22.71} \\
\midrule
\textbf{$M = 5$} \\
NES-RE & 93.8 $\pm$ 0.0 & \textbf{0.030 $\pm$ 0.001} & -1165 $\pm$ 5.503 & 74.3 $\pm$ 0.2 & 0.071 $\pm$ 0.004 & -5134 $\pm$ 60.72 \\
BQ-S & 93.7 $\pm$ 0.1 & 0.032 $\pm$ 0.000 & \textbf{-1113 $\pm$ 4.123} & 74.5 $\pm$ 0.1 & \textbf{0.055 $\pm$ 0.002} & \textbf{-4897 $\pm$ 25.66} \\
\midrule
\textbf{$M = 10$} \\
NES-RE & 93.8 $\pm$ 0.0 & \textbf{0.030 $\pm$ 0.001} & -1159 $\pm$ 5.959 & 74.3 $\pm$ 0.2 & 0.069 $\pm$ 0.004 & -5083 $\pm$ 58.42 \\
BQ-S & 93.8 $\pm$ 0.0 & 0.031 $\pm$ 0.000 & \textbf{-1098 $\pm$ 3.752} & \textbf{74.7 $\pm$ 0.1} & \textbf{0.045 $\pm$ 0.001} & \textbf{-4766 $\pm$ 15.89} \\
\bottomrule
\end{tabular}
}
\caption{Test accuracy, expected calibration error (ECE), and log likelihood (LL) on CIFAR-10 and CIFAR-100 for BQ-S (our proposal) and NES-RE (the strongest baseline) for the ``Slimmable Network'' search space.
We see that BQ-S consistently outperforms NES-RE in terms of ECE and LL, whilst maintaining the same accuracy.}
\label{tab:supernet}
\end{table}

Finally, we perform experiments to examine robustness to dataset shift.
Previous work has provided evidence that ensembling of Neural Neworks provides robustness to shifts in the underlying data distribution \citep{zaidi_neural_2022,shu_neural_2022}.
However, these investigations have assumed the availability of a validation set from the shifted distribution, which we argue is unrealistic in practice.
Instead, we examine the setting where only the test set is shifted, and the validation set is representative of the training set.
We use the benchmark established by \citet{hendrycks_benchmarking_2019} to generate shifted datasets by applying one of 30 corruption types to each image for CIFAR-10 and CIFAR-100.
Each corruption type has a severity level on a $1-5$ scale.
Table \ref{tab:robustness} shows a comparison between NES-RE and BQ-S in this setting (on the slimmable network search space).
We see that, whilst our proposal performs similarly in terms of accuracy, it produces ensembles that perform significantly better in terms of expected calibration error and test set log likelihood.
This trend holds across corruption severity levels.

\begin{table}[t]
\vspace{-4mm}
\centering
\resizebox{0.95\linewidth}{!}{%
\begin{tabular}{lccccccc}
\toprule \\
& \multicolumn{6}{c}{Severity Level 1} \\
& \multicolumn{3}{c}{CIFAR-10} & \multicolumn{3}{c}{CIFAR-100} \\
\cmidrule(lr){2-4} \cmidrule(lr){5-7}
Algorithm & Accuracy & ECE & LL & Accuracy & ECE & LL \\
\midrule
\textbf{$M = 3$} \\
NES-RE & 86.20 $\pm$ 0.04 & 0.046 $\pm$ 0.001 & -59259.6 $\pm$ 595.907 & 62.36 $\pm$ 0.08 & 0.151 $\pm$ 0.004 & -169235 $\pm$ 1632.43166 \\
BQ-S & 86.26 $\pm$ 0.08 & \textbf{0.036 $\pm$ 0.001} & \textbf{-54283.4 $\pm$ 642.383} & 62.52 $\pm$ 0.10 & \textbf{0.093 $\pm$ 0.002} & \textbf{-149022 $\pm$ 364.57480} \\
\midrule
\textbf{$M = 5$} \\
NES-RE & 86.25 $\pm$ 0.04 & 0.046 $\pm$ 0.001 & -59178.3 $\pm$ 851.719 & 62.36 $\pm$ 0.09 & 0.155 $\pm$ 0.003 & -169999 $\pm$ 1553.96433 \\
BQ-S & 86.16 $\pm$ 0.06 & \textbf{0.032 $\pm$ 0.001} & \textbf{-52173.6 $\pm$ 202.350} & \textbf{62.58 $\pm$ 0.09} & \textbf{0.103 $\pm$ 0.004} & \textbf{-152466 $\pm$ 1249.96873} \\
\midrule
\textbf{$M = 10$} \\
NES-RE & 86.26 $\pm$ 0.04 & 0.043 $\pm$ 0.001 & -57010.4 $\pm$ 722.311 & 62.46 $\pm$ 0.07 & 0.145 $\pm$ 0.002 & -164816 $\pm$ 975.78975 \\
BQ-S & 86.22 $\pm$ 0.05 & \textbf{0.029 $\pm$ 0.001} & \textbf{-50504.6 $\pm$ 443.984} & 62.54 $\pm$ 0.08 & \textbf{0.093 $\pm$ 0.002} & \textbf{-149022 $\pm$ 364.57480} \\
\bottomrule \\
& \multicolumn{6}{c}{Severity Level 3} \\
& \multicolumn{3}{c}{CIFAR-10} & \multicolumn{3}{c}{CIFAR-100} \\
\cmidrule(lr){2-4} \cmidrule(lr){5-7}
Algorithm & Accuracy & ECE & LL & Accuracy & ECE & LL \\
\midrule
\textbf{$M = 3$} \\
NES-RE & 73.16 $\pm$ 0.08 & 0.147 $\pm$ 0.002 & -133205 $\pm$ 1537.15 & 49.18 $\pm$ 0.07 & 0.235 $\pm$ 0.005 & -270710 $\pm$ 2628.75462 \\
BQ-S & 73.31 $\pm$ 0.12 & \textbf{0.131 $\pm$ 0.002} & \textbf{-123113 $\pm$ 1498.55} & \textbf{49.45 $\pm$ 0.12} & \textbf{0.193 $\pm$ 0.007} & \textbf{-250337 $\pm$ 3304.95} \\
\midrule
\textbf{$M = 5$} \\
NES-RE & 73.18 $\pm$ 0.09 & 0.148 $\pm$ 0.002 & -133239 $\pm$ 1904.50 & 49.20 $\pm$ 0.09 & 0.239 $\pm$ 0.004 & -272004 $\pm$ 2482.09961 \\
BQ-S & 73.23 $\pm$ 0.07 & \textbf{0.125 $\pm$ 0.001} & \textbf{-118756 $\pm$ 614.899} & \textbf{49.57 $\pm$ 0.09} & \textbf{0.175 $\pm$ 0.005} & \textbf{-241407 $\pm$ 2438.78} \\
\midrule
\textbf{$M = 10$} \\
NES-RE & 73.23 $\pm$ 0.08 & 0.143 $\pm$ 0.002 & -128663 $\pm$ 1664.07 & 49.29 $\pm$ 0.07 & 0.227 $\pm$ 0.003 & -263639 $\pm$ 1596.01214 \\
BQ-S & 73.39 $\pm$ 0.11 & \textbf{0.120 $\pm$ 0.002} & \textbf{-114613 $\pm$ 1247.44} & \textbf{49.57 $\pm$ 0.06} & \textbf{0.163 $\pm$ 0.003} & \textbf{-235152 $\pm$ 881.120} \\
\bottomrule \\
& \multicolumn{6}{c}{Severity Level 5} \\
& \multicolumn{3}{c}{CIFAR-10} & \multicolumn{3}{c}{CIFAR-100} \\
\cmidrule(lr){2-4} \cmidrule(lr){5-7}
Algorithm & Accuracy & ECE & LL & Accuracy & ECE & LL \\
\midrule
\textbf{$M = 3$} \\
NES-RE & 55.49 $\pm$ 0.08 & 0.285 $\pm$ 0.002 & -239927 $\pm$ 2187.24 & 34.04 $\pm$ 0.06 & 0.339 $\pm$ 0.005 & -415182 $\pm$ 3710.25 \\
BQ-S & 55.67 $\pm$ 0.14 & \textbf{0.265 $\pm$ 0.003} & \textbf{-226433 $\pm$ 2523.79} & \textbf{34.24 $\pm$ 0.11} & \textbf{0.291 $\pm$ 0.008} & \textbf{-385063 $\pm$ 5355.88} \\
\midrule
\textbf{$M = 5$} \\
NES-RE & 55.53 $\pm$ 0.08 & 0.286 $\pm$ 0.003 & -240154 $\pm$ 2835.89 & 34.04 $\pm$ 0.07 & 0.344 $\pm$ 0.005 & -417110 $\pm$ 3476.95 \\
BQ-S & 55.51 $\pm$ 0.05 & \textbf{0.260 $\pm$ 0.002} & -\textbf{220196 $\pm$ 1198.20} & \textbf{34.35 $\pm$ 0.10} & \textbf{0.270 $\pm$ 0.006} & \textbf{-371575 $\pm$ 4083.67} \\
\midrule
\textbf{$M = 10$} \\
NES-RE & 55.54 $\pm$ 0.08 & 0.279 $\pm$ 0.002 & -233441 $\pm$ 2474.23 & 34.11 $\pm$ 0.07 & 0.331 $\pm$ 0.003 & -405068 $\pm$ 2327.88 \\
BQ-S & 55.61 $\pm$ 0.11 & \textbf{0.254 $\pm$ 0.003} & \textbf{-214126 $\pm$ 2030.20} & \textbf{34.35 $\pm$ 0.07} & \textbf{0.257 $\pm$ 0.003} & \textbf{-361876 $\pm$ 1666.07} \\
\bottomrule
\end{tabular}
}
\caption{Test accuracy, expected calibration error (ECE), and log likelihood (LL) on CIFAR-10 and CIFAR-100 for NES-RE (the strongest baseline), and BQ-S (our strongest proposal) using the ``Slimmable Network'' search space for a range of corruption severities.
We see that BQ-S is more robust than NES-RE to dataset shift, especially in terms of LL and ECE.}
\label{tab:robustness}
\end{table}

\section{Discussion and Future Work}
We proposed a method for building ensembles of Neural Networks using the tools provided by Bayesian Quadrature.
Specifically, by viewing ensembling as approximately performing marginalisation over architectures, we used the warped Bayesian Quadrature framework to select a candidate set of architectures to train.
We then suggest two methods of constructing the ensemble based upon this candidate set: one based upon recombination of the approximate posterior over architectures (BQ-R), and one based upon optimisation of the ensemble weights (BQ-S) using a validation set.
BQ-R approximately performs hierarchical Bayesian inference using BQ, whereas BQ-S is a heuristic inspired by BQ.
The discrepancy in performance is likely due to the fact that BQ-R does not make use of the validation set, as it takes the Bayesian perspective and performs hierarchical inference over both architecture weights and architectures using the training set.
(In principle, BQ-R can use the union of the training and validation sets to perform hierarchical inference. However, we did not run experiments in this setting as it would obviously allow BQ-R significantly more compute than the alternative methods.)
BQ-S (and all the baselines), however, make use of a separate validation set to select the ensemble weights.
We additionally show that BQ-S outperforms state-of-the-art baselines when the search space is large, and on the largest datasets for smaller search spaces.
This is likely because it is more exploratory than alternative methods, and so less likely to become stuck near local minima of the architecture likelihood.
Lastly, we demonstrated that BQ-S is more robust to dataset shift than state-of-the-art baselines.

An interesting direction for future work is to examine the effect of marginalising over architecture weights as well as over architectures.

We introduce a general-purpose method, so its societal impacts will depends on the specific tasks to which it is applied. We find it difficult to anticipate what those tasks will be, and even more difficult to speculate meaningfully about any societal impacts will be.

\subsubsection*{Acknowledgments}
The authors would like to thank (withheld for anonymisation).

\bibliography{references}
\bibliographystyle{tmlr}

\end{document}